\documentclass[11pt]{article}

\usepackage[margin=1in]{geometry}
\usepackage[utf8]{inputenc}
\usepackage[T1]{fontenc}
\usepackage{lmodern}
\usepackage{microtype}
\usepackage{url}
\usepackage{booktabs}
\usepackage{amsfonts}
\usepackage{amsmath}
\usepackage{nicefrac}
\usepackage{xcolor}
\usepackage{graphicx}
\usepackage{pgfplots}
\pgfplotsset{compat=1.17}
\usepackage[round]{natbib}
\usepackage[colorlinks=true,linkcolor=blue!60!black,citecolor=blue!60!black,urlcolor=blue!60!black]{hyperref}

\title{Choosing the Number of Attention Heads from Intrinsic Dimension:\\
A Small-Scale Study}

\author{
\small
\begin{tabular}{cc}
Arjun Boya & Charan Sai Bandaru \\
Institute of Cognitive Science & Institute of Cognitive Science \\
University of Osnabr\"uck & University of Osnabr\"uck \\
\texttt{aboya@uni-osnabrueck.de} &
\texttt{cbandaru@uni-osnabrueck.de} \\[1.5em]
Masthanaiah Pillimetla & Kavana Chidanandappa \\
Department of Social Sciences and Technology &
Department of Social Sciences and Technology \\
Technical University of Munich &
Technical University of Munich \\
\texttt{masthanaiahpillimetla7@gmail.com} &
\texttt{ckavana003@gmail.com}
\end{tabular}
}

\date{}

\begin{document}

\maketitle

\begin{abstract}
The number of attention heads in a transformer is usually set by convention or grid search. We test a simple heuristic for setting it in advance: estimate the intrinsic dimension of token representations without labels, and use roughly that many heads. We estimate intrinsic dimension with a nested-dropout VAE trained on DistilBERT embeddings of WikiText-2, choosing the KL weight at the knee of the VAE's rate--distortion curve rather than by hand. This gives $k^*=9.2$ (95\% CI $[8.6, 9.8]$). In a small 4-layer transformer trained from scratch, 10 heads and 32 heads are statistically equivalent on SST-2 and on MRPC (paired TOST over 10 seeds, $\pm1$ point margin), with 69\% fewer attention parameters for the 10-head model. Training 10 heads from the start also slightly outperforms training 32 heads and pruning to 10 (0.6 points, $p=0.008$). The evidence is narrow: two classification tasks, one small architecture, no pretrained models, and no formal link between intrinsic dimension and head count. We present the result as a practical heuristic worth testing at larger scale, not as a general principle.
\end{abstract}

\section{Introduction}

Ask why a transformer has 8, 12 or 16 attention heads per layer and the honest answer is usually that a previous model did. Head count is rarely derived from the task. Meanwhile, pruning studies keep finding that many heads in trained models can be removed with little loss \citep{michel2019sixteen, voita2019analyzing}, which suggests the usual counts are often larger than needed. The standard response is to train a large model and prune it afterwards. We ask whether the count can instead be chosen before training, from a cheap property of the data.

The property we use is intrinsic dimension: roughly, the number of independent directions along which the representations actually vary. Section~\ref{sec:motivation} explains why this might bear on head count and where the analogy breaks down. In short, we treat it as a hypothesis with a testable prediction: a model with about $k^*$ heads should do as well as one with many more.

We test that prediction at small scale. Our contributions are:
\begin{itemize}
    \item A label-free procedure for estimating a head count: a nested-dropout VAE on token embeddings, an elbow detector suited to the convex spectra such VAEs produce, and a rule for choosing the VAE's KL weight from its own rate--distortion curve. Applied to DistilBERT embeddings of WikiText-2, it gives $k^*=9.2$.
    \item A paired-seed comparison showing that 10 heads match 32 heads on SST-2 and MRPC (TOST equivalence within $\pm1$ accuracy point) in a 4-layer transformer trained from scratch, with 69\% fewer attention parameters.
    \item A comparison with reactive pruning: training 10 heads from the start slightly but consistently beats training 32 heads and pruning to 10.
\end{itemize}

The claims are modest. We do not derive head count from intrinsic dimension, and two small tasks with one architecture do not establish a general rule. Section~\ref{sec:limitations} lists what would be needed to do so.

\section{Related Work}

\paragraph{Head redundancy and pruning.} \citet{michel2019sixteen} showed that most heads in trained transformers can be removed at test time with small accuracy loss, using an importance score based on the effect of masking each head. \citet{voita2019analyzing} found that a few specialized heads do most of the work and pruned the rest during training. Both start from a fixed, large head count and reduce it. We instead choose the count before training and compare against this kind of pruning directly.

\paragraph{Intrinsic dimension of representations.} Neural representations typically occupy far fewer dimensions than their ambient size. \citet{ansuini2019intrinsic} measured this across layers of vision networks with the TwoNN estimator \citep{facco2017estimating}, and \citet{aghajanyan2021intrinsic} showed that fine-tuning language models needs only a low-dimensional subspace of parameter updates. These works use intrinsic dimension to describe trained models; we try to use it to configure one.

\paragraph{Ordered latent variables.} Nested dropout \citep{rippel2014nested} trains an autoencoder so that its latent dimensions are ordered by importance, which makes it possible to read off how many dimensions are in use from a single model. We combine it with the variational autoencoder \citep{kingma2014autoencoding} and treat the KL weight, as in $\beta$-VAE \citep{higgins2017betavae}, as a rate--distortion knob. For locating elbows we start from the Kneedle method \citep{satopaa2011kneedle}.

\section{Why Intrinsic Dimension Might Bear on Head Count}
\label{sec:motivation}

Each attention head projects tokens into its own small subspace (16 dimensions in our models) and computes a separate attention pattern there. More heads let a layer track more token-to-token relations in parallel, at a parameter cost that grows linearly with the head count. The question is how many of these parallel views a task can actually use.

Our working hypothesis is that this number is limited by how many independent directions of variation the token representations have. If the embeddings vary along about $k$ independent factors, a layer with far more than $k$ heads has little new to attend to with the extra ones, and they should end up redundant or unused. That would be consistent with how much trained models can be pruned. Under this hypothesis, an estimate $k^*$ of intrinsic dimension suggests a head count, and setting the count near $k^*$ should cost little accuracy compared with a much larger default.

This is an analogy, not a derivation, and it has obvious gaps. A head is not tied to a single latent dimension. Deeper layers see transformed representations, not the input embeddings. And nothing guarantees that $k^*$ is a tight bound rather than a loose one. We therefore treat the hypothesis only as a source of a testable prediction: about $k^*$ heads should match a much larger count. Its practical appeal is that $k^*$ can be estimated once per domain, without labels and before any classifier is trained.

\section{Method}
\label{sec:method}

\subsection{Estimating $k^*$ with a nested-dropout VAE}
\label{sec:vae}

\paragraph{Data.} We pass passages from the WikiText-2 training split \citep{merity2017pointer} through a frozen, pretrained \texttt{distilbert-base-uncased} \citep{sanh2019distilbert} and take the final-layer hidden states, skipping lines shorter than 20 characters, truncating at 128 tokens and dropping the \texttt{[CLS]} and \texttt{[SEP]} positions. We keep the first 60,000 token vectors (768-dimensional) and standardize each dimension. These embeddings come from neither downstream task, so the estimate does not depend on SST-2 or MRPC.

\paragraph{Model.} The VAE encoder maps each 768-dimensional vector through layers of 512 and 256 units (each with LayerNorm and GELU) to the mean and log-variance of a 64-dimensional latent code; the decoder mirrors it. During training, nested dropout \citep{rippel2014nested} samples $m$ uniformly from $\{1,\dots,64\}$ at each step and zeroes latent dimensions $m{+}1$ to $64$. Early dimensions are therefore always in use and later ones rarely, so the trained model orders its dimensions by importance. Training details are in Appendix~\ref{app:hparams}.

\paragraph{Reading off a dimension.} After training, we compute the variance of each latent mean across a held-out 10\% of the embeddings and sort these variances in decreasing order. The resulting spectrum drops steeply and then flattens. We take its elbow as the number of active dimensions: after normalizing both axes to $[0,1]$, the elbow is the point furthest \emph{below} the straight line joining the first and last points. Because the spectrum is convex, this is the correct side of the line; measuring distance above it, as the concave-curve version of Kneedle does, returns an endpoint instead. Appendix~\ref{app:elbow} shows this with synthetic spectra.

\paragraph{Choosing the KL weight.} The elbow depends strongly on the KL weight $\beta$: stronger regularization switches off more latent dimensions. Rather than picking $\beta$ by hand, we train the VAE at ten values, $\beta \in \{0.1, 0.25, 0.5, 1, 2, 3, 4, 5, 6, 8\}$, with five seeds each, and record the held-out reconstruction loss. Plotting mean loss against $\beta$ traces a rate--distortion curve, and we take $\beta^*$ at its knee, found with the same line-distance rule (both axes min--max normalized, $\beta$ on a linear scale). The estimate $k^*$ is the mean elbow across seeds at $\beta^*$. Nothing in this procedure looks at the downstream tasks.

\subsection{Downstream comparison: 10 heads versus 32}
\label{sec:method-heads}

\paragraph{Tasks.} SST-2 \citep{socher2013recursive} sentiment classification (67,349 training and 872 validation sentences) and MRPC \citep{dolan2005automatically} paraphrase detection from GLUE \citep{wang2019glue} (3,668 training and 408 validation sentence pairs). Text is tokenized with the \texttt{bert-base-uncased} tokenizer to at most 128 tokens; MRPC pairs are encoded jointly as \texttt{[CLS]} sentence 1 \texttt{[SEP]} sentence 2 \texttt{[SEP]}.

\paragraph{Model.} A 4-layer transformer encoder \citep{vaswani2017attention} trained from scratch, with model dimension 128, feed-forward dimension 512 and 16 dimensions per head. Classification uses the \texttt{[CLS]} position. Only the number of heads $k$ differs between configurations, so the attention projection width is $16k$: 160 for $k=10$ and 512 for $k=32$. The two models have 4,788,418 and 5,509,314 parameters in total, and 327,680 and 1,048,576 attention parameters, a 69\% reduction in attention parameters for $k=10$.

\paragraph{Design.} We train both configurations with the same 10 random seeds. For a given seed, both see identical data order (each run has its own seeded data loader) and differ only in head count. We test the paired accuracy differences in two ways: a two-sided paired $t$-test for a difference, and two one-sided tests (TOST, \citealp{lakens2017equivalence}) for equivalence within $\pm1$ accuracy point. We fixed this margin before running the comparison. $k=10$ was fixed before these experiments (Appendix~\ref{app:provenance} gives its history) and was used unchanged on MRPC, which therefore serves as an out-of-sample check. MRPC needed different optimization settings from SST-2 because it is about 18 times smaller (Appendix~\ref{app:mrpc}); both head counts always share the same settings.

\subsection{Pruning baseline}
\label{sec:method-pruning}

To compare prospective selection with the usual alternative, for each of the same 10 seeds we train a 32-head model on SST-2, score each head by how much masking it increases validation loss \citep{michel2019sixteen}, keep the 10 highest-scoring heads in each layer, and build a genuinely smaller model by slicing out their weights. We checked that this sliced model reproduces the masked 32-head model up to floating-point rounding (largest difference below $10^{-6}$ over five random trials). The pruned model is then fine-tuned for up to 5 epochs at a learning rate of $2\times10^{-5}$.

\section{Results}
\label{sec:results}

\subsection{The VAE estimate}
\label{sec:results-vae}

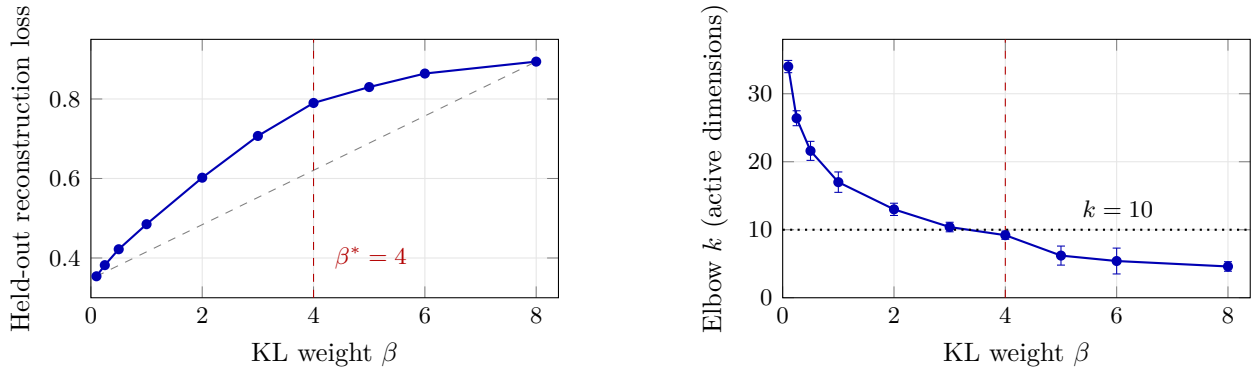
\begin{figure}[t]
\centering
\begin{tikzpicture}
\begin{axis}[
  width=0.47\linewidth, height=5cm,
  xlabel={KL weight $\beta$}, ylabel={Held-out reconstruction loss},
  xmin=0, xmax=8.4, ymin=0.3, ymax=0.95,
  tick label style={font=\footnotesize}, label style={font=\small},
  grid=major, grid style={gray!20},
]
\addplot[mark=*, mark size=1.5pt, blue!70!black, thick] coordinates {
  (0.1,0.354) (0.25,0.382) (0.5,0.422) (1,0.485) (2,0.602)
  (3,0.707) (4,0.790) (5,0.830) (6,0.864) (8,0.894)};
\addplot[dashed, gray] coordinates {(0.1,0.354) (8,0.894)};
\addplot[dashed, red!70!black] coordinates {(4,0.3) (4,0.95)};
\node[font=\footnotesize, red!70!black, anchor=west] at (axis cs:4.2,0.40) {$\beta^*=4$};
\end{axis}
\end{tikzpicture}
\hfill
\begin{tikzpicture}
\begin{axis}[
  width=0.47\linewidth, height=5cm,
  xlabel={KL weight $\beta$}, ylabel={Elbow $k$ (active dimensions)},
  xmin=0, xmax=8.4, ymin=0, ymax=38,
  tick label style={font=\footnotesize}, label style={font=\small},
  grid=major, grid style={gray!20},
]
\addplot[mark=*, mark size=1.5pt, blue!70!black, thick,
  error bars/.cd, y dir=both, y explicit] coordinates {
  (0.1,34.0) +- (0,0.9) (0.25,26.4) +- (0,1.1) (0.5,21.6) +- (0,1.4)
  (1,17.0) +- (0,1.5) (2,13.0) +- (0,0.9) (3,10.4) +- (0,0.7)
  (4,9.2) +- (0,0.6) (5,6.2) +- (0,1.4) (6,5.4) +- (0,1.9) (8,4.6) +- (0,0.7)};
\addplot[dashed, red!70!black] coordinates {(4,0) (4,38)};
\addplot[dotted, black, thick] coordinates {(0,10) (8.4,10)};
\node[font=\footnotesize, anchor=south west] at (axis cs:5.2,10.3) {$k=10$};
\end{axis}
\end{tikzpicture}
\caption{The $\beta$ sweep (mean over 5 seeds). Left: held-out reconstruction loss; $\beta^*$ is the point furthest from the dashed chord. Right: elbow of the latent variance spectrum, with 95\% confidence intervals. At $\beta^*=4$ the elbow is $k^*=9.2$; the dotted line marks the head count tested downstream.}
\label{fig:rd}
\end{figure}

Figure~\ref{fig:rd} and Table~\ref{tab:vae} show the sweep. Held-out reconstruction loss rises steadily with $\beta$ while the elbow falls from 34 active dimensions at $\beta=0.1$ to about 5 at $\beta=8$, so the estimate depends heavily on the KL weight and a single hand-picked value would be close to arbitrary. The knee of the loss curve is at $\beta^*=4$, an interior point (its distance from the chord peaks there at 0.222 and falls to zero at both ends). At $\beta^*$ the elbow is $k^*=9.2$, 95\% CI $[8.6, 9.8]$. The values 3, 5 and 6 were added after a first seven-point sweep to check the knee more finely; they left $\beta^*$ and $k^*$ unchanged.

The head count we test, $k=10$, sits just above this interval, 0.8 dimensions from the point estimate. We read this as rough agreement, not a precise match.

\begin{table}[t]
  \caption{Values behind Figure~\ref{fig:rd}: held-out reconstruction loss and elbow $k$ (mean $\pm$ 95\% CI half-width over 5 seeds). Italic rows were added after the first seven-point sweep.}
  \label{tab:vae}
  \centering
  \small
  \begin{tabular}{lcc}
    \toprule
    $\beta$ & Reconstruction loss & Elbow $k$ \\
    \midrule
    0.10 & 0.354 & $34.0 \pm 0.9$ \\
    0.25 & 0.382 & $26.4 \pm 1.1$ \\
    0.50 & 0.422 & $21.6 \pm 1.4$ \\
    1.00 & 0.485 & $17.0 \pm 1.5$ \\
    2.00 & 0.602 & $13.0 \pm 0.9$ \\
    \textit{3.00} & \textit{0.707} & \textit{$10.4 \pm 0.7$} \\
    \textbf{4.00} ($\beta^*$) & \textbf{0.790} & $\mathbf{9.2 \pm 0.6}$ \\
    \textit{5.00} & \textit{0.830} & \textit{$6.2 \pm 1.4$} \\
    \textit{6.00} & \textit{0.864} & \textit{$5.4 \pm 1.9$} \\
    8.00 & 0.894 & $4.6 \pm 0.7$ \\
    \bottomrule
  \end{tabular}
\end{table}

\subsection{10 heads versus 32}
\label{sec:results-heads}

Table~\ref{tab:cross-task} gives the main result. On SST-2 the 10-head model averages 82.13\% validation accuracy and the 32-head model 81.95\%. The paired difference is not significant ($t=1.08$, $p=0.31$, Cohen's $d=0.34$), and TOST confirms equivalence within one point ($p=0.0005$). On MRPC, with $k=10$ carried over unchanged, the two models reach 70.47\% and 70.37\% ($t=0.34$, $p=0.74$, $d=0.11$), again equivalent ($p=0.006$). So in this setting, cutting attention parameters by 69\% costs no measurable accuracy on either task.

\begin{table}[t]
  \caption{Validation accuracy of 10 versus 32 heads (mean $\pm$ std over 10 paired seeds). TOST tests equivalence within $\pm1$ accuracy point.}
  \label{tab:cross-task}
  \centering
  \small
  \begin{tabular}{lcccc}
    \toprule
    Task & $k=10$ & $k=32$ & Paired $t$-test & TOST \\
    \midrule
    SST-2 & $82.13\% \pm 0.42$ & $81.95\% \pm 0.33$ & $p=0.31$ & equivalent, $p=0.0005$ \\
    MRPC  & $70.47\% \pm 0.90$ & $70.37\% \pm 1.03$ & $p=0.74$ & equivalent, $p=0.006$ \\
    \bottomrule
  \end{tabular}
\end{table}

MRPC accuracy is low for this task, and both models overfit heavily (about 94\% training accuracy against 70\% validation), which is what one expects when training a transformer from scratch on 3,668 examples. The gap is the same for both head counts (23.70 and 23.69 points), so it does not bias the comparison, but the absolute MRPC numbers say little about pretrained models.

\subsection{Comparison with pruning}
\label{sec:results-pruning}

Pruning the trained 32-head models to 10 heads without further training drops accuracy sharply and erratically, to $71.69\% \pm 7.04$ (Table~\ref{tab:pruning}). Fine-tuning recovers most of this, to $81.50\% \pm 0.37$. The 10-head models trained from scratch still do slightly better, by 0.63 points on average ($t=3.41$, $p=0.008$, $d=1.08$). The gap is consistent across seeds but small; it also falls inside our $\pm1$ point equivalence margin (TOST $p=0.039$). The larger difference is cost. The prospective model never needs a trained 32-head model or a head-scoring pass; the full pruning pipeline took 2.41 GPU-hours for 10 seeds, about as much as all 20 runs of the SST-2 comparison (Appendix~\ref{app:repro}).

The comparison is if anything tilted toward pruning, because head scores and fine-tuning both used the validation set on which accuracy is reported. We tested only one importance score, on one task.

\begin{table}[t]
  \caption{Pruning baseline on SST-2 (mean $\pm$ std over the same 10 seeds).}
  \label{tab:pruning}
  \centering
  \small
  \begin{tabular}{lc}
    \toprule
    Configuration & Validation accuracy \\
    \midrule
    32 heads, unpruned & $81.95\% \pm 0.33$ \\
    32 pruned to 10, before fine-tuning & $71.69\% \pm 7.04$ \\
    32 pruned to 10, after fine-tuning & $81.50\% \pm 0.37$ \\
    10 heads from the start & $\mathbf{82.13\% \pm 0.42}$ \\
    \bottomrule
  \end{tabular}
\end{table}

\section{Discussion and Limitations}
\label{sec:limitations}

What the experiments support is narrow: for this small model on these two tasks, 10 heads are enough, and choosing 10 up front is at least as good as pruning down to 10. The VAE estimate landing near 10 is encouraging but is a single agreement, not a validation of the method. Several things would be needed before treating the heuristic as reliable.

\paragraph{No formal link.} Section~\ref{sec:motivation} gives an intuition, not a theorem. We have no bound connecting intrinsic dimension to the number of heads a task needs, and no account of why a property of the input embeddings should constrain every layer.

\paragraph{Too few head counts.} We compare only 10 and 32. That shows 10 is sufficient, not that $k^*$ is the right place to stop. A sweep over roughly 4 to 16 heads with multiple seeds would show where accuracy starts to fall and whether that point tracks $k^*$. The same sweep would answer an obvious objection: that a simpler rule, such as trying a few powers of two, would find an equally good count without any VAE.

\paragraph{One estimator.} We use one intrinsic-dimension estimator. Others, such as TwoNN \citep{facco2017estimating} or PCA explained variance, might give different numbers, and the knee rule for $\beta$ is one reasonable choice among several (cross-validated likelihood or minimum description length would be alternatives).

\paragraph{Small scale.} All results come from a 4-layer, 128-dimensional transformer trained from scratch on two binary classification tasks. We have not tested pretrained models, larger models, generation, long contexts or tasks that need multi-step reasoning, where head requirements may be quite different.

\paragraph{Narrow pruning baseline.} One ablation-based importance score on one task. Gradient-based scores \citep{michel2019sixteen} or pruning during training \citep{voita2019analyzing} might close the gap.

\section{Conclusion}

We tested whether a label-free estimate of intrinsic dimension can set the number of attention heads before training. In a small transformer, a head count near the estimate matched one more than three times larger on two tasks and slightly beat pruning to the same size. These results are a reason to try the heuristic at a larger scale, with a proper sweep over head counts and comparisons against simpler rules, not a reason to adopt it yet.

\section*{Reproducibility}

Code, the per-seed result files and the scripts behind every table, including the synthetic elbow test and the pruning-slicing check, are available at \url{https://github.com/subcortical7/IT-MHA}. Hyperparameters are listed in Appendix~\ref{app:hparams}.

\appendix

\section{Development History}
\label{app:history}

Several problems surfaced while building this pipeline and changed its results. We record them here because they explain some design choices in the main text.

\subsection{Unpaired seeds}
\label{app:pairing}

In an early version of the SST-2 comparison, both head counts drew data order from the global random number generator after model initialization. Initialization consumes a number of random draws that depends on the head count, so for the same seed the 10-head and 32-head runs saw different batch orders, and the paired test was not actually paired. Every run now has its own seeded data loader, and we confirmed that batch order matches within each seed pair before rerunning all comparisons.

\subsection{Elbow detection on convex spectra}
\label{app:elbow}

The first elbow detector measured distance \emph{above} the chord joining the spectrum's endpoints, which suits concave curves. A variance spectrum from nested dropout is convex, so every interior point lies below the chord and the detector returns an endpoint (or, with noise, a point next to one), chosen by noise rather than by the curve's shape. We tested this on synthetic spectra. On 15 noiseless convex spectra the original rule returned $k=1$ or $k=64$ every time, and on 260 noisy ones it returned $k \in \{1, 2, 63, 64\}$ every time. The corrected rule, which measures distance below the chord, recovered the known elbow exactly on all 180 noisy piecewise-linear spectra with elbows between $k=4$ and $k=32$.

\subsection{Where $k=10$ came from}
\label{app:provenance}

$k=10$ was first obtained from an early version of the VAE analysis that used the faulty elbow detector at a hand-picked $\beta=4$. That number should not be trusted on its own, which is why we later replaced the hand-picked $\beta$ with the knee rule and fixed the detector. The corrected procedure chooses $\beta^*$ without reference to the earlier result, happens to land on the same $\beta^*=4$, and gives $k^*=9.2$. So $k=10$ was fixed before the downstream experiments, but the corrected estimate that supports it came afterwards. We regard the agreement as a modest corroboration, not an independent derivation.

\subsection{MRPC training settings}
\label{app:mrpc}

Reusing the SST-2 settings on MRPC made both models predict only the majority class (validation accuracy stuck at 68.4\%). The cause was the fixed 500-step warmup, which at batch size 64 covered most of MRPC's 580--870 training steps. A first fix with weaker regularization (dropout 0.15, weight decay 0.01) avoided the collapse but let both models overfit, and the 32-head model overfit slightly more: its train--validation gap was about 0.7 points larger. Under that setting the paired difference reached $p=0.04$ in favor of 10 heads. We did not treat this as a real head-count effect, since it came from unequal overfitting. The final settings (Appendix~\ref{app:hparams}) equalize the gaps and the difference disappears. This suggests head-count comparisons on small datasets should be checked for regularization-driven asymmetries before a difference is reported.

\section{Hyperparameters}
\label{app:hparams}

\begin{table}[h]
  \caption{Training settings. All classifier runs use AdamW, a cosine learning-rate schedule with linear warmup, gradient clipping at 1.0, and early stopping on validation accuracy.}
  \label{tab:hparams}
  \centering
  \small
  \begin{tabular}{lcccc}
    \toprule
    & SST-2 & MRPC & Pruned fine-tune & VAE \\
    \midrule
    Learning rate & $10^{-4}$ & $2\times10^{-4}$ & $2\times10^{-5}$ & $10^{-3}$ \\
    Weight decay & 0.05 & 0.03 & 0.05 & $10^{-4}$ \\
    Dropout & 0.30 & 0.25 & 0.30 & -- \\
    Batch size & 64 & 32 & 64 & 256 \\
    Max epochs & 10 & 20 & 5 & 40 \\
    Early-stopping patience & 3 & 6 & 2 & -- \\
    Warmup & 500 steps & 6\% of steps & 5\% of steps & -- \\
    KL annealing & -- & -- & -- & $0\to\beta$ over 20 epochs \\
    Seeds & 10 & 10 & 10 & 5 per $\beta$ \\
    \bottomrule
  \end{tabular}
\end{table}

Classifier seeds are 42, 123, 456, 789, 1011, 2024, 31337, 7, 999 and 8675309; VAE seeds are 42, 123, 7, 2024 and 99. The VAE is trained on a random 90\% of the embeddings and evaluated on the remaining 10\%.

\section{Compute}
\label{app:repro}

All experiments ran on a single NVIDIA GPU workstation. Recorded training time was about 5.2 GPU-hours: 2.55\,h for the 20 SST-2 runs, 0.24\,h for the 20 MRPC runs and 2.41\,h for the pruning baseline (training the 32-head models, scoring heads and fine-tuning). The VAE sweep was not timed.

\end{document}